%% file: neurips_2023.tex
\documentclass{article}

\usepackage[nonatbib, final]{neurips_2023_ml4ps}

\usepackage[utf8]{inputenc} 
\usepackage[style=numeric,sorting=none]{biblatex}
\usepackage[T1]{fontenc}    
\usepackage{hyperref}       
\usepackage{url}            
\usepackage{booktabs}       
\usepackage{amsfonts}       
\usepackage{nicefrac}       
\usepackage{microtype}      
\usepackage{xcolor}         

\title{Weak neural variational inference for solving Bayesian inverse problems {\em without} forward models: applications in elastography}

\author{Vincent C.~Scholz$^a$, Yaohua Zang$^a$, Phaedon-Stelios Koutsourelakis$^{a,b}$ \\
$^a$ Technical University of Munich, Professorship of Data-driven Materials Modeling,  \\
School of Engineering and Design, Boltzmannstr. 15, Garching, Germany \\
$^b$ Munich Data Science Institute (MDSI - Core member), Garching, Germany \\
\texttt{\{vincent.scholz, yaohua.zang, p.s.koutsourelakis\}@tum.de}\\
}
    
\include{definitions}

\begin{document}

\maketitle

\begin{abstract}
In this paper, we introduce a novel, data-driven approach for solving high-dimensional Bayesian inverse problems based on partial differential equations (PDEs), called Weak Neural Variational Inference (WNVI).
The method complements real measurements with virtual observations derived from the physical model. In particular,  weighted residuals are employed as probes to the governing PDE in order to formulate and  solve a Bayesian inverse problem \textit{without} ever formulating nor solving a forward model. 
The formulation treats the state variables of the physical model as latent variables, inferred using Stochastic Variational Inference (SVI), along with the usual unknowns. The approximate posterior employed uses neural networks to approximate the inverse mapping from state variables to the unknowns.
We illustrate the proposed method in a biomedical setting where we infer spatially varying material properties from noisy tissue deformation data. We demonstrate that WNVI is not only as accurate and more efficient  than traditional methods that rely on repeatedly solving the (non)linear forward problem as a black-box, but it  
can  also handle ill-posed forward problems (e.g., with insufficient boundary conditions). 
\end{abstract}


\section{Introduction}
\input{Sections/1_Introduction}

\newpage
\section{Problem definition} \label{sec:prob_definition}
\input{Sections/2_Problem_definition}
\section{Proposed method / Methodology} \label{sec:method}
\input{Sections/3_Methodology}

\section{Numerical Illustrations} \label{sec:num_exp}
\input{Sections/4_NumericalExperiments}

\section{Conclusions} \label{sec:outlook}
\input{Sections/5_Outlook}

\section*{Acknowledgement}
The support of the Deutsche
Forschungsgemeinschaft (DFG) through Project Number  499746055 is gratefully acknowledged.

\clearpage
\printbibliography

\appendix
\input{Sections/App_Integration}
\input{Sections/App_ELBO}
\input{Sections/App_NN}
\input{Sections/App_Jump_penalty}
\input{Sections/App_ComparisonDetails}

\end{document}

%% file: definitions.tex
\usepackage[dvips]{graphicx}
\usepackage{amsmath}
\usepackage{amssymb}
\usepackage{psfrag}
\usepackage{hyperref}

\usepackage{subfigure}

\usepackage{xcolor}			
\usepackage{color}			

\usepackage[normalem]{ulem}

\usepackage{tikz}

\usepackage{setspace}
\usepackage{algorithm}
\usepackage{algpseudocode}

\newcommand{\bs}{\boldsymbol}

\newcommand{\refeq}[1]{Equation (\ref{#1})}

\newcommand{\ee}{\end{equation}}
\newcommand{\be}{\begin{equation}}
\newcommand{\ec}{\end{center}}
\newcommand{\bc}{\begin{center}}
\newcommand{\eea}{\end{eqnarray}}
\newcommand{\bea}{\begin{eqnarray}}
\newcommand{\bd}{\begin{description}}
\newcommand{\ed}{\end{description}}
\newcommand{\bi}{\begin{itemize}}
\newcommand{\ei}{\end{itemize}}

\newcommand{\bx}{\bs{x}}
\newcommand{\bxi}{\bx^{(i)}}

\newcommand{\by}{\bs{y}}

\newcommand{\bt}{\bs{\theta}}

\newcommand{\ve}[1]{\boldsymbol{#1}} 

\newcommand{\person}[1]{\textsc{#1}}
\newcommand{\diff}[1]{\mathrm{d} #1}

\usepackage{etoolbox}
\pretocmd{\eqref}{Eq.~}{}{}

\usepackage{multirow}

%% file: Sections/1_Introduction.tex
Model-based inverse problems enhance our ability to extract information from data by enabling the identification  of unknown model parameters or functions. In this manner, one can gain unparalleled insight into the state of the  system of interest as well as calibrate associated models to obtain accurate predictions about the system's evolution. 
The solution of model-based inverse  problems 
has had significant impact in several fields such as medical imaging \cite{kak2001principles}, climate modeling \cite{jackson2004efficient}, geophysics \cite{hill2006effective}, and astronomy \cite{craig1986inverse}.
One such application is model-based elastography \cite{doyley2012model}, a technique employed in medical imaging to identify the mechanical properties of biological tissue based on tissue deformation data. The latter are obtained with, e.g., ultrasound or MRI and arise from static or time-dependent excitations. 
The material  properties are identified in a non-intrusive manner and provide a characterization of the tissue, which medical practitioners can use to diagnose patients and, in some cases, detect the appearance of pathologies earlier as compared to classical imaging techniques alone (e.g., x-rays, MRI) \cite{ganne-carrie_accuracy_2006, sayed_breast_2020, hoyt_tissue_2008, asbach_assessment_2008, hamhaber_vivo_2010, ohayon_biomechanics_2014, shore_transversely_2011, schmitt_noninvasive_2007}. 
In this case, the model consists of the (non)linear Partial Differential Equations (PDEs) of solid mechanics with associated boundary/initial conditions.
These  are subsequently discretized, and the resulting system of thousands or millions of (non)linear (and potentially time-dependent) algebraic  equations constitutes the so-called forward model.
While these are commonplace in the engineering mechanics community, the discretization and use of such models is out-of-reach for the majority of medical practitioners. This has been a significant obstacle in the permeation 
of such techniques despite the wealth of useful information they can furnish. It should also be noted that in several circumstances, the availability of a forward model is not guaranteed. In the simplest case where mild pressure is applied, e.g., on a patient's abdomen with an ultrasound wand, which is used to acquire images of the deformation in the underlying tissue, boundary conditions are in part  or in whole, missing \cite{barbone2004elastic, mclaughlin2003unique}. Irrespective of the aforementioned issues and even in cases where a well-posed forward model is available, the solution of the inverse problem requires multiple solutions and derivatives of the model-predicted displacements/strains with respect to the sought parameters. As each call to the forward model solver can be  cumbersome,  the solution of the inverse problem might be slow and require significant investments in dedicated computing hardware.

The aforementioned difficulties are to various degrees also present in the entirety of PDE-based inverse problems. In combination with the unavoidable noise in the observation data \cite{barbone2002quantitative}
 contribute to the well-documented {\em ill-posedness} of inverse problems \cite{tarantola2005inverse}. As our formulation is Bayesian in nature we primarily review pertinent techniques \cite{kaipio2006statistical}. Such formulations do not yield point estimates but rather quantify the probability with which the unknowns take any of the possible values. Bayesian estimates can be of great importance in medical diagnostics in particular as they can guide decision-making and inform the need for more data or  tests. 
Classical probabilistic inference techniques such as \person{Markov}-chain Monte-Carlo (MCMC) \cite{green2015bayesian}, Sequential Monte Carlo (SMC) \cite{Koutsourelakis2009} and Variational Inference (VI) \cite{blei2017variational} exhibit very poor scaling in terms of the number of forward-model calls with the dimension of the vector of unknowns \cite{roberts1996exponential, mattingly2012diffusion}. As a result, several efforts have been directed toward reducing the intrinsic dimension \cite{spantini_inference_2018,franck2016sparse} of the unknowns or reducing the cost of the each  forward-model solve.
In the latter case, a popular direction involves the development of surrogates for the forward map based on, e.g.   polynomial chaos \cite{Li2014, Marzouk2007}, \person{Gauss}ian processes \cite{Chen2020, Bilionis2013}, reduced-order models \cite{sheriffdeen_accelerating_2019}, deep neural networks \cite{Li2019, Zhu2018, mo2019deep}, neural operators  \cite{li2020fourier,lu2019deeponet,kaltenbach2023semi}
which are frequently employed in multi-fidelity schemes \cite{Peherstorfer2018, Yan2019, nitzler2020generalized}. It is noted that such surrogates have also been proposed for the inverse map \cite{padmanabha2021solving,kaltenbach2023semi}. Generally the surrogates are trained in an offline-phase using several input-output pairs which imply an equal number of forward-model solves. Apart from the cost associated with data generation which increases rather fast with the input dimension, the trained surrogate can introduce bias or additional (epistemic) uncertainty which should be accounted during inference \cite{Bilionis2013Solution}. More importantly perhaps and given that the posterior is generally concentrated on a minute and a-priori unknown  region of the parameter space, it becomes questionable whether the surrogate can provide sufficient accuracy in that  region. 

Physics-informed machine learning tools have more recently been proposed such as PINNs \cite{raissi2019physics}, weak adversarial networks \cite{bao2020numerical, zang2020weak}), DeepONets \cite{lu2019deeponet,kaltenbach2023semi}, PINO  \cite{li2021physics}. These methods employ collocation residuals of the governing PDE in the training loss and as such can produce solutions to the inverse problem without even solving the forward problem. When the unknowns consist of a spatially-varying property field as in the case of elastography, they express this with a neural network which complicates the incorporation of prior information when this is available. As a result, they have difficulties capturing jumps in the underlying property field as is the case when inclusions-tumors are present in the tissue \cite{krishnapriyan2021characterizing}. In \cite{yang2021b, liu2023bayesian}, a  Bayesian  version of PINNs  was proposed where a  posterior on the weights/biases of the neural network expressing the unknown field is inferred. Although only collocation residuals (and their derivatives) need to be evaluated at each iteration, their total number, and as a result the total computational cost, can be comparable to classical techniques due to the local information they provide and the lack of quantification of their informational content.

In this paper, we build upon previous work \cite{koutsourelakis2012novel, bruder2018beyond} that attempts to overcome the tyranny of an expensive black-box solver for the forward model.  We propose a probabilistic learning objective in which weighted residuals of the governing PDE are used as {\em virtual} observables, which are combined with the actual observables via a {\em virtual} likelihood \cite{kaltenbach2020incorporating, rixner2021probabilistic}.
{Most similar to our method  is perhaps the work of \cite{vadeboncoeur2023fully} who employ all the weighted residuals and a different architecture for synthesising actual observables as well as for inferring the unknown parameters. 
We show that selecting a few weighted residuals at random at each iteration offers a theoretically consistent formulation  that is sufficient to infer the unknown, spatially varying property field and to obtain probabilistic estimates without ever solving the forward model \cite{zhu_physics-constrained_2019}. Our contributions can be summarized as follows:
\begin{itemize}
\item we obviate the need for a well-posed forward problem. As such, the formulation can be employed in cases where, e.g., boundary conditions are unknown or insufficiently specified.

\item as a further consequence, inference of the posterior does not require any costly forward model evaluations nor adjoint computations for the evaluation of derivatives of the (log-)likelihood. In addition, it eliminates the need for cumbersome  discretizations (e.g. finite differences/elements) of the governing equations.

\item updates require the evaluation of a reduced number of  weighted residuals (and their derivatives), i.e., integrals over the problem domain,  which accelerates inference.

\item as a result, the solution of inverse problems based on linear or non-linear forward models proceeds identically without the need for methodological adjustments, nor does it imply any additional computational burden.

\end{itemize}
The subsequent sections of this paper are structured as follows: In Section \ref{sec:prob_definition}, we introduce the forward and \person{Bayes}ian inverse problem for elastography. Section \ref{sec:method} outlines our general methodology and derivation of the evidence lower bound (ELBO) necessary for approximating the posterior as well as the form of approximate posterior. In Section \ref{sec:num_exp}, we demonstrate the efficiency of our approach through numerical experiments. Finally, Section \ref{sec:outlook} concludes with a summary and outlook.

%% file: Sections/2_Problem_definition.tex
In this section, we discuss the main aspects of the problem of model-based elastography \cite{oberai2009linear}. In particular, the governing PDEs that constitute the forward model as well as the data based on which one can formulate the Bayesian inverse problem. We note that the essential elements and the resulting challenges are identical for most other PDE-based inverse problems.
As mentioned in the introduction, the governing equations arise from solid mechanics and, in particular, the conservation of linear momentum, which,  in the case of time-independent settings and the absence of body forces, dictates that: 
\begin{equation}
    \label{eqn:PDE}
     \nabla \cdot \ve \sigma(\ve s) =  0 \quad\quad ~\bs{s} \in  \Omega \subset \mathbb{R}^d,
\end{equation}
where $\ve \sigma$ 
is the \person{Cauchy} stress tensor that is a function of space $\bs{s} \in \Omega$,
 and a point-wise (local)  constitutive  law which we express as: 
\begin{equation}
    \label{eqn:const_law}
    \ve \sigma(\ve s) = \ve \sigma(\nabla \ve u(\ve s) ;  ~m(\ve s)).
\end{equation}
where $\ve u(\bs{s})$ denotes the displacement field  and $m(\bs{s})$  the material parameters (e.g., Young's modulus), which generally vary in space \footnote{In order to simplify the  notation, we will omit indicating explicitly  dependence on space $\ve s$ when it is obvious from the context.}.
We remark that most of the numerical experiments are carried out under the assumption of small displacements/deformations, in which case the stress tensor $\ve \sigma$ depends on $ \ve u$ through the infinitesimal strain tensor $\ve \epsilon =\frac{1}{2}(\nabla \ve u + (\nabla \ve u)^T)$. 
This dependence, i.e., the constitutive law above, can be linear or nonlinear.

A well-posed forward model  necessitates the prescription of boundary conditions, either of \person{Dirichlet} or  \person{Neumann} type, which can be expressed as follows: 
\begin{align}
     \ve u & = \ve f  \quad\quad \textrm{ on } \Gamma_\mathrm{D}, \\
     \ve \sigma \cdot \vec{\ve n} & = \ve g \quad\quad \textrm{ on } \ \Gamma_\mathrm{N}, \label{eqn:bcs}
\end{align}
where $\ve f$ and $\ve g$ are given 
 on the boundaries $\Gamma_\mathrm{D}$ and $\Gamma_\mathrm{N}$ (with $\Gamma_\mathrm{D} \cap \Gamma_\mathrm{N} = \emptyset$,  $\Gamma_\mathrm{D} \cup \Gamma_\mathrm{N} = \partial\Omega$) and $\vec{\ve n}$ is the outward, unit normal vector.
 
We emphasize that, as in all PDE-based inverse problems, the model is assumed to be correct, up at least to any discretization errors. This assumption is especially precarious given the phenomenology of constitutive laws and the variability of biological tissues. It constitutes a potential source of {\em model error/bias} \cite{koutsourelakis2012novel} which,  if present, does not prevent  one from solving the  inverse problem  and obtaining estimates of the material parameters $m$, but, as  it is easily understood,  these estimates would be erroneous or even misleading. 

We denote abstractly with $\ve u(m)$ the displacement field arising by solving the aforementioned equations and its dependence on the material property field $m$. 
In the context of the inverse elastography problem, one is supplied with measurements $\hat{\ve u}=\{\hat{u}_i\}_{i=1}^{N_{\hat{\ve u}}}$ of the  displacements  obtained usually through ultrasound or MRI at specific locations of the domain $\Omega$. 
Each such measurement $\hat{ u}_i$ is usually assumed to relate to the model-predicted displacement $ u_i(m)$ at the same point as:
\begin{equation}
	\hat{ u}_i =  u_i(m) + 
 \tau^{-0.5}  \epsilon_i, \qquad \epsilon_i \sim \mathcal{N}(\ve 0, \ve 1),
 \label{eq:uhatnoise}
\end{equation}
where the additive term accounts for the stochastic observation noise, which is assumed to be Gaussian distributed with variance $\tau^{-1}$. 
In the probabilistic (i.e., Bayesian) framework advocated in this work, the aforementioned equation gives rise to a likelihood $p(\hat{\ve u}| m)$ which, when combined with a prior $p(m)$, gives rise to the sought posterior density $p(m | \hat{\ve u})$
\footnote{To simplify the presentation and minimize notational complexity, we have not considered at this stage the discretization of the governing PDE nor of the unknown field $m$. This does not imply that prior/posterior densities on spatially varying fields must be specified, which requires appropriate mathematical care.}:

\be
\begin{array}{ll}
p(m | \hat{\ve u}) &  \propto p(\hat{\ve u}| m) ~p(m) \\
 & \propto \prod_{i=1}^{N_{\hat{\ve u}}} \mathcal{N} (\hat{\ve u}_i~|~\ve u_i(m), ~\tau^{-1}) ~~p(m).
\end{array}
\label{eqn:posterior_unnormed} 
\ee
Classical inference techniques employ the PDE-solver as a black box, which implies that a solution to the PDE is required for each likelihood evaluation. Especially when the representation of the unknown field $m$ is high-dimensional, derivatives of the (log)posterior/likelihood are  required to explore the search space efficiently. This, in turn, necessitates a differentiable PDE-solver, which can be achieved using adjoint formulations \cite{oberai_evaluation_2004}. This adds further complication, which, as mentioned in the introduction, creates an insurmountable modeling barrier for medical practitioners.  We further note that differentiability is not available in most  legacy solvers,  which would preclude their use in the context of such inverse problems.  We emphasize that the requirement for derivatives is also present when deterministic solutions to the inverse problem are sought \cite{oberai_linear_2009}. 

%% file: Sections/3_Methodology.tex
\subsection{Overview of the framework} \label{subsec:overview}

We propose a reformulation of the inverse problem that obviates the need for a forward model while retaining the valuable information that the governing PDE provides. 
We view the latter as an information source, which we probe with weighted residuals. Before embarking on the presentation of the particulars, and to simplify subsequent derivations, we postulate a finite-dimensional representation of the displacement field $\bs{u}$ (i.e., the state variable in the governing PDE) and the unknown material parameter field $m$, which takes the form:
\begin{equation}
     m(\ve x, \ve s) = \sum_{i=1}^{d_{\ve x}} x_i \eta^{\ve x}_i(\ve s) \quad \mathrm{\ and \ } \quad \ve u(\ve y, \ve s) = \sum_{i=1}^{d_{\ve y}} y_i \bs{\eta}^{\ve y}_i(\ve s),
     \label{eq:discr}
\end{equation}
where $\eta^{\ve x}_i$ and $\bs{\eta}^{\ve y}_i$ are given feature functions dependent on space and $\bx=\{x_i\}_{i=1}^{d_{\ve x}}$ , $\by=\{y_i\}_{i=1}^{d_{\ve y}}$ the corresponding coefficients, respectively. The former can be, e.g., the usual Finite Element shape functions, radial basis functions, spectral representations based on, e.g., sines/cosines, wavelets, Chebyshev polynomials \cite{fanaskov_spectral_2023,vadeboncoeur2023fully}.
Their role is {\em not} to provide a discretization scheme for the governing PDE but merely a representation of these fields. As a result, they do not need to be associated with any, e.g., finite-element mesh as those typically employed to solve the forward problem. In general, $d_{\ve x}, d_{\ve y}>>1$ to capture the full details. One could also employ neural networks (NNs) to represent these fields. In this case, the analogs of the $\bx,\by$ above would be the associated weights/biases.

We consider weighted residuals of the governing PDE in Equations (\ref{eqn:PDE})-(\ref{eqn:bcs}) based on vector-valued  weight functions $\ve w(\ve s) \in \mathcal{W} \subset H^1(\Omega)$ such that $\ve w|_{\Gamma_D}=0$.  By employing integration-by-parts, one arrives at the following expression  (in indicial notation) for each such weight function $\bs{w}$ \cite{belytschko2014nonlinear}:
\begin{equation}
    \label{eqn:residual}    
    r_{\bs{w}}(\bx,\by) = \int_{\Omega} \sigma_{ij} w_{i,j} \diff\Omega - \int_{\Gamma_N}  g_i w_i \diff\Gamma_N, 
\end{equation}
which implicitly depends on $\bx$ and $\by$ given \refeq{eq:discr} and the dependence of the stresses $\ve \sigma$ on $\ve u$ through the constitutive equation in (\ref{eqn:const_law}). 
We note that depending on  the choice of the weight functions $\ve w$, one can obtain a minimum of six methods (i.e., collocation, sub-domain, least-squares, (Petrov)-Galerkin, moments) \cite{finlayson2013method}.
As it will become evident, the weighted residuals serve as data sources rather than as a means to deriving a discretized system of equations as it is done conventionally. 
Consequently, we have the flexibility to explore alternative or even non-symmetric versions concerning $\ve u$, $\ve \sigma$, or $\ve w$. For instance, this includes expressions involving lower-order derivatives of the candidate solution $\ve u$, obtained through further applications of integration by parts \cite{kharazmi2021hp}.

We consider a set of $N$ distinct weight functions $\bs{w}^{(j)}$ and corresponding weighted residuals $r_{\ve w^{(j)}}(\bx,\by)$. We assume that a {\em virtual} observation $\hat{r}_j=0$ for each residual is available, which relates to the actual residual as:
\be
0=\hat{r}_j= r_{\ve w^{(j)} }(\by,\bx)+\lambda^{-1}\epsilon_j, \qquad \epsilon_j \sim \mathcal{N}(0,1)
\ee
If we denote summarily all the {\em virtual} observables  $\hat{\bs{R}}=\{ \hat{r}_j=0 \}_{j=1}^N$, the equation above gives rise to a {\em virtual} likelihood: 
\be
\begin{array}{ll}
 p(\hat{\bs{R}} | \by, \bx) & =\prod_{j=1}^N p(\hat{r}_j=0 | ~\by, \bx) \\
 & \propto \prod_{j=1}^N \sqrt{\lambda}  \exp\left(  - \frac{\lambda}{2} ~r_{\ve w^{(j)}}^2(\by,\bx) \right),
  \end{array}
\label{eq:virtuallike}
\ee
where the hyper-parameter  $\lambda>0$ penalizes the deviation of the weighted residuals from $0$, i.e. the value they would attain for any solution pair $\bx,\by$ (or equivalently $m$ and $\ve u$). Selecting values for $\lambda^{-1}$ similar to the numerical tolerance of a deterministic iterative solver has proven to be a good strategy as its role is similar in nature. 
We note that evaluating the \textit{virtual} likelihood  involves computing $N$ integrals over the problem domain (see \refeq{eqn:residual}) as compared to the \textit{black-box} forward solver, which classical formulations entail.
The integrations over the problem domain can either be carried out deterministically or by Monte Carlo (see Appendix \ref{app:integration}).

The {\em virtual} likelihood above is complemented by the actual likelihood of the displacement observations $\hat{\ve u}$. In particular, by evaluating the displacement field in \refeq{eq:discr} at the locations where displacements are provided, we obtain $ u_i(\by)$ which as in \refeq{eqn:posterior_unnormed} give rise to the {\em actual} likelihood:
\be
\begin{array}{ll}
p(\hat{\ve u}|\by)  & =\prod_{i=1}^{N_{\hat{\ve u}}} \mathcal{N} (\hat{ u}_i~|~ u_i( \by), ~\tau^{-1}) \\
& \propto \prod_{i=1}^{N_{\hat{\ve u}}} \sqrt{ \tau}  e^{-\frac{\tau}{2} (\hat{ u}_i-u_i( \by))^2}
\end{array}
\label{actual_like}
\ee

A mere application of Bayes rule would then yield the following {\em joint posterior} on $\bx$ and $\by$:
\be
\begin{array}{ll}
p(\bx,\by | \hat{\ve R}, \hat{\ve u}) = \cfrac{p(\hat{\ve u}|\by) p(\hat{\bs{R}} | \by, \bx)~p(\by,\bx)}{p(\hat{\ve R}, \hat{\ve u})}
\end{array}
\label{eq:jointposterior}
\ee
where $p(\by,\bx)$ denotes the prior and $p(\hat{\ve R}, \hat{\ve u})$ the model evidence.
With regards to the former, in subsequent numerical illustrations,  we  employ the following form:
\begin{equation}
    p(\ve y, \ve x) = p(\ve y ) p(\ve x),
\end{equation}
where $p(\ve x)$ may be the same prior as in a traditional formulation. A vague, uninformative prior was employed for $ p(\ve y )$, reflecting prior beliefs about the PDE solution as represented by $\by$.
 Naturally, other forms could be employed, including hyper-parameters, which could be fine-tuned during inference  it is done in \cite{vadeboncoeur2023fully}. In our formulation however, the role of $p(\ve x,  \ve y)$ is not to provide accurate predictions of the forward or backward map between  $\by$ and  $\bx$, as this is the purpose of the posterior, which is discussed in the subsequent section.

\noindent \textbf{Remarks}
\bi
\item In contrast to classical Bayesian formulations, our sought posterior is defined jointly on $\bx$ and $\by$, which must both be inferred. The increased dimension of the state space allows us to circumvent the black-box solver. In fact, the formulation and solution of the inverse problem \textbf{does not require the availability of a forward model or even a well-posed forward problem}. As we show in the sequel, the form of the terms involved on the right-hand side of \refeq{eq:jointposterior} enables fast inference despite the increased dimension.

\begin{figure}
    \centering
    \includegraphics{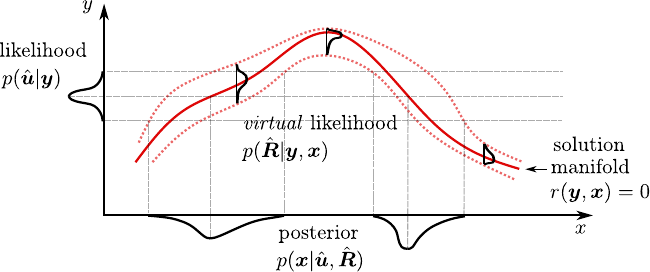}
    \caption{Schematic illustration of the proposed formulation. The actual data $\hat{\ve u}$ induces an actual likelihood $p(\hat{\ve u}| \ve y)$ on the (representation of the) PDE-solution $\by$. This is combined with the {\em virtual} likelihood $p(\hat{\ve R}| \ve y, \ve x)$ that assigns probabilities to $(\bx,\by)$ pairs. The virtual likelihood is highest on the solution manifold, i.e., for all $(\bx,\by)$ pairs that yield $0$ values for the weighted residuals $r(\by,\bx)$ and decays at a rate determined by the precision $\lambda$ as one moves away from it.  The combination of these two likelihoods (and the priors which are not depicted here) gives rise to the sought posterior $p(\ve x |\hat{\ve u}, \hat{\ve R})$ on (the representation of) the unknown material field $\bx$.
    }
    \label{fig:virtual_likelihood}
\end{figure}

\item The posterior obtained will depend on the virtual observables $\hat{\bs{R}}$ employed both in terms of their number $N$ as well as in terms of the weight functions these are based on. In general, the higher the $N$, the more information from the PDE is incorporated and the closer one would be to the solution manifold that consists of all pairs of $\bx$ and $\by$ that yield $0$ residuals, i.e., they are solution pairs of the PDE (Figure \ref{fig:virtual_likelihood}).

\ei

\subsection{Probabilistic Inference and randomization of residuals} \label{subsec:ELBO}
The exact posterior in \refeq{eq:jointposterior} is generally intractable. For that purpose, we advocate the use of Variational Inference, i.e., we seek an approximation from a parametrized family of densities $q_{\ve \xi}(\ve x, \ve y)$ by minimizing the \person{Kullback-Leibler} (KL) divergence with the exact posterior. If $\bs{\xi}$ denotes the tunable parameters, then this is equivalent to maximizing the Evidence Lower BOund (ELBO) $\mathcal{L}(\bs{\xi})$ to the log-evidence $\log p(\hat{\ve R}, \hat{\ve u})$ \cite{blei2017variational}:
\be
\begin{array}{ll}
    \log p(\hat{\ve R}, \hat{\ve u}) & = \log \int p(\hat{\ve u} | \ve y) p(\hat{\ve R} | \ve y, \ve x) p(\ve y , \ve x) \diff\ve y \diff\ve x \\
    &\geq \left< \log \frac{p(\hat{\ve u} | \ve y) p(\hat{\ve R} | \ve y, \ve x) p(\ve y , \ve x)}{q_{\ve \xi}(\ve y, \ve x)} \right>_{q_{\ve \xi}(\ve y, \ve x)} \\ 
    
    & = -\frac{\lambda}{2} \sum_{j=1}^N \left< r^2_{\ve w^{(j)}} (\ve y, \ve x) \right>_{q_{\ve \xi}(\ve y, \ve x)} -\frac{\tau}{2}  \sum_{i=1}^{N_{\hat{\ve u}}} \left< (\hat{ u}_i -  u_i(\by))^2 \right>_{q_{\ve \xi}(\ve y, \ve x)} \\
    & + \left< \log \frac{p(\ve y , \ve x)}{q_{\ve \xi}(\ve y, \ve x)} \right>_{q_{\ve \xi}(\ve y, \ve x)} \\
    & = \mathcal{L}(\ve \xi),
\end{array}
\label{eqn:ELBO}
\ee
where $\left< \cdot \right>_{q_{\bs{\xi}}}$ denotes the expectation with respect to $q_{\bs{\xi}}$. The first term of the ELBO promotes the (on average) minimization of the $N$ weighted residuals, the second minimizes (on average) the  discrepancy between model predictions and  the measurements, whereas the third provides a regularization by minimizing the KL-divergence with the prior. We note the resemblance of the ELBO with alternative formulations that have been proposed, such as  PINNs \cite{raissi2019physics} or weak adversarial networks \cite{bao2020numerical}. Generally, these employed ad-hoc coefficients as relative weights for the aforementioned terms and made use of  collocation-type residuals, which arise as a special case in our formulation (i.e., by using Dirac-deltas as the weight functions). More importantly, however, our formulation enables a fully probabilistic interpretation in which the information content of the weighted residuals can be quantified, as discussed in the sequel.

In \refeq{eqn:ELBO}, it is evident that maximizing the ELBO requires repeated evaluations of the $N$ weighted residuals (as well as their derivatives with respect to the parameters $\ve \xi$). As mentioned earlier, the higher the $N$ is, the more information from the governing PDE is incorporated, but the higher the computational cost becomes. 
To improve the efficiency of this computation,  we advocate a Monte Carlo approximation for the first term in \refeq{eqn:ELBO}, i.e.:
\begin{align}
    \sum_{j=1}^N \left< r^2_{\ve w^{(j)}} (\ve y, \ve x) \right>_{q_{\ve \xi}(\ve y, \ve x)} \approx & \frac{N}{K} \sum_{k=1}^K \left< r^2_{\ve w^{(j_k)}} (\ve y, \ve x) \right>_{q_{\ve \xi}(\ve y, \ve x)}, \qquad j_k \sim Cat\left(N, \frac{1}{N} \right). 
    \label{eqn:approx_virt_like}
\end{align}
Thus, we randomly sample $K << N$ weight functions and the corresponding $K$ weighted residuals provide an unbiased Monte Carlo estimator of this  ELBO term even for $K=1$.

In section \ref{sec:elboalg} and in greater detail in Appendix Appendix \ref{app:ELBO}, we explain how to compute the gradients of $\mathcal{L}$ and use them for its  maximization. We also note that if a parameterized prior, e.g. $p_{\bs{\zeta}}(\bx, \by)$ were used, the optimal $\bs{\zeta}$ can be found by maximizing the resulting ELBO \cite{rixner2021probabilistic}. While these structured or informative priors could greatly improve inference speed and prediction accuracy, this paper does not explore their potential.

We finally note that the ELBO provides a quantitative metric of the information content of each weighted residual. As a result, and rather than randomly sub-sampling residuals, one could envision procedures that could lead to selecting weighting functions of superior informational content, which are not explored in this paper. Another possibility could be to localize weight functions in regions of the problem domain where inferential uncertainty is higher or which are of increased interest to the analyst. Similar to the way one can zoom in on a photo, the framework advocated employs weight functions as indirect magnifying glasses that enable one to zoom in on the underlying and unobserved material property field.


\subsection[Approximation of q]{Approximate posterior $q_{\boldsymbol{\xi}}$} \label{subsec:approx_posterior}
The form of the approximate posterior   $q_{\bs{\xi}}$ plays a crucial  role in the efficacy and accuracy of the formulation. It has to be expressive enough to approximate  the actual posterior and as simple as possible to increase  computational efficiency and interpretability.
It is clear that $\bx$ and $\by$ are strongly dependent as they must jointly  ensure that the weighted residuals in the virtual likelihood are in the vicinity of zero. Capturing this dependence is crucial to properly quantifying the uncertainty in $\bx$, i.e., the representation of the material property field $m$. To this end we factorize the joint, approximate posterior $q_{\bs{\xi}}$ as follows:
\be
q_{\bs{\xi}}(\bx,\by)=q_{\bs{\xi}}(\bx |\by)~q_{\bs{\xi}}(\by)
\label{eq:qjoint}
\ee
The first of these densities takes the form:
\be
    \label{eqn:q_x}
    q_{\bs{\xi}}(\ve x | \ve y) = \mathcal{N}\left( \ve x | ~\bs{\mu}_{x;\xi_x}\left( \ve y \right), ~\bs{S}_x 
    \right),
\end{equation}
where the conditional mean $\bs{\mu}_x$ is parametrized by a neural network with tunable parameters $\bs{\xi}_x$,  the details of which are contained in Appendix \ref{app:NN}.
The conditional covariance $\bs{S}_x$ is assumed to be independent of $\by$ and of the form:
\be
\bs{S}_x=\ve L_x \ve L_x^T + \mathrm{diag}\left( \ve \sigma_x^2 \right)
\ee
where $\bs{L}_x$ is a matrix of dimension $d_{\ve x} \times d_{\tilde{\ve x}}$ that captures the principal directions along which (conditional) variance is larger whereas  $\bs{\sigma}_x^2$ is a vector of dimension $d_x$ that captures the residual (conditional) variance along the $\bx-$ dimensions.
In contrast to a full covariance matrix, the form adopted for $\bs{S}_x$  ensures linear scaling of the unknown parameters with $d_x$, which for most problems can be high.

With regards to the second density in \refeq{eq:qjoint}, we adopt the following form:
\be
  q_{\bs{\xi}}(\ve y) =\mathcal{N}(\by |~ \ve \mu_y, \ve S_y)
  \label{eq:qy}
 \ee 
 where again a similar form for $\bs{S}_y$ is adopted in order to ensure linear scaling with $d_y$, i.e.:
 \begin{equation}
    \label{eqn:Sigma_y}
    \ve S_y = \ve L_y \ve L_y^T + \mathrm{diag}\left( \ve \sigma_y^2 \right),
\end{equation}
where $\ve L_y$ is rectangular matrix of dimensions $d_{\ve y} \times d_{\tilde{\ve y}}$ (where $d_{\tilde{\ve y}} \ll d_{\ve y}$) and $\ve \sigma_y^2$ is a vector of dimensions $d_{\ve y}$.
As a result the vector of parameters $\bs{\xi}$ that are optimized in order to maximize the ELBO $\mathcal{L}(\bs{\xi})$ consists of:
\be
\bs{\xi}=\{ \bs{\xi}_x, \bs{L}_x, \bs{\sigma}_x^2, \ve \mu_y ,\bs{L}_y, \bs{\sigma}_y^2 \}
\ee
and is of dimension $dim(\bs{\xi}_x)+d_{\ve x} \times (d_{\tilde{\ve x}} + 1) + d_{\ve y} \times (d_{\tilde{\ve y}} + 2)$.
Algorithmic details about the maximization of the ELBO are contained in section \ref{sec:elboalg}. In section \ref{sec:probest}, we describe how the density $q_{\bs{\xi}}$ can be used to obtain probabilistic estimates of the unknown material property field $m$.


\subsection{Stochastic Variational Inference for the maximization of the ELBO}
\label{sec:elboalg}
The maximization of the ELBO $L(\bs{\xi})$ is carried out in the context of Stochastic Variational Inference \cite{hoffman2013stochastic}, which is based on Monte Carlo estimates of $L(\bs{\xi})$ and its gradient in conjunction with Stochastic Gradient Ascent for the maximization.
Regarding the former, we use the reparameterization trick \cite{kingma2013auto} to generate samples from the approximate posterior $q_{\ve \xi}$. 
Based on the form of $q_{\bs{\xi}}$ discussed in the previous section, this can be done by drawing first $\by$-samples from $q_{\bs{\xi}}(\by)$ (\refeq{eq:qy}) as: 
\begin{equation}
    \by=\ve \mu_y + \bs{L}_y \ve \varepsilon_1+ \ve \sigma_y  \odot \ve \varepsilon_2, \quad  \ve \varepsilon_1 \sim \mathcal{N}\left(\ve 0, \ve I_{d_{\tilde{\ve y}}}\right) \mathrm{ \ and \ } \ve \varepsilon_2 \sim \mathcal{N}\left(\ve 0, \ve I_{d_{\ve y}}\right)
\label{eq:ysamp}
\end{equation}
and in a second step using them to sample $\bx$ from the conditional posterior $q_{\bs{\xi}}(\bx|\by)$ (\refeq{eqn:q_x}) as:
\be
\bx=\bs{\mu}_{x;\xi_x}\left( \ve y \right) +\ve L_x \ve \varepsilon_3  +  \ve \sigma_x \odot  \ve \varepsilon_4, \quad  \ve \varepsilon_3 \sim \mathcal{N}\left(\ve 0, \ve I_{d_{\tilde{\ve x}}}\right) \mathrm{ \ and \ } \ve \varepsilon_4 \sim \mathcal{N}\left(\ve 0, \ve I_{d_{\ve x}}\right)
\label{eq:xsamp}
\ee
where $\bs{\mu}_{x;\xi_x}(.)$ denotes the  neural network described in Appendix \ref{app:NN}.
We generate $L$ such sample pairs $( \ve x ,  \ve y )$ 
 which are used to produce Monte Carlo estimates of the expectations with respect to $q_{\bs{\xi}}$ appearing in the ELBO (\refeq{eqn:ELBO}) and its gradient,  
 which is computed via PyTorch's automatic differentiation capability \cite{NEURIPS2019_9015}. 
 Detailed expressions  can be found in Appendix \ref{app:ELBO}, and a pseudo-code is shown in Algorithm \ref{alg:SVI}.

\begin{algorithm}[t]
\caption{SVI Training Algorithm}\label{alg:SVI}
\begin{algorithmic}
\State Select $\lambda$, $\tau$, $K$, $L$; Initialize $\ve \xi \leftarrow \ve \xi_0$, $t \leftarrow 0$
\While{$\mathcal{L}$ not converged}
\State Generate $K$ weight functions $\bs{w}^{(j_k)}$ 
\For{$\ell=1$ to  $L$}
\State Draw $\ve y_\ell \leftarrow \ve \mu_y + \bs{L}_y \ve \varepsilon_1 + \ve \sigma_y \ve \varepsilon_2,\ \ve \varepsilon_1 \sim \mathcal{N}\left(\ve 0, \ve I_{d_{\tilde{\ve y}}}\right), \ve \varepsilon_2 \sim \mathcal{N}\left(\ve 0, \ve I_{d_{\ve y}}\right)$ 
\State $\bs{\mu}_{n,x} \leftarrow NN(\ve y_\ell)$ \Comment{see \refeq{eqn:q_x}}
\State Draw $\ve x_\ell \leftarrow \bs{\mu}_{x;\xi_x}\left( \ve y_\ell \right) + \bs{L}_x \ve \varepsilon_3 + \ve \sigma_x \ve \varepsilon_4, \  \ve \varepsilon_3 \sim \mathcal{N}\left(\ve 0, \ve I_{d_{\tilde{\ve x}}}\right), \ve \varepsilon_4 \sim \mathcal{N}\left(\ve 0, \ve I_{d_{\ve x}}\right)$ 
\EndFor
\State Estimate $\mathcal{L}_{\ve \xi}(\ve x_\ell, \ve y_\ell, w^{(j_k)})$ \Comment{see \refeq{eqn:app_full_elbo}}
\State Estimate $\nabla_{\ve \xi} \mathcal{L}_{\ve \xi}$ \Comment{see Appendix \ref{app:ELBO}}
\State Update $\ve \xi_{t + 1} \leftarrow \ve \xi_{t} + \ve \rho^{(t)} \odot \nabla_{\ve \xi} \mathcal{L}_{\ve \xi}$ \Comment{Step sizes $\ve \rho$ via ADAM}
\State $t \leftarrow t + 1$
\EndWhile 
\end{algorithmic}
\end{algorithm}


\subsection{Probabilistic estimates of unknown material property field}
\label{sec:probest}
After convergence of the SVI scheme and upon identification of the optimal values for $\bs{\xi}$, probabilistic estimates of the  unknown material property field $m(\bs{s})$ through the approximate posterior $q_{\bs{\xi}}(\ve y, \ve x)$.
In particular, posterior predictive samples $m_{b}(\ve s)$   of the material property field at arbitrary locations $\bs{s} \in \Omega$ can be obtained  by drawing  samples, say $\ve x_{b}$, from  $q_{\bs{\xi}}(\ve y, \ve x)$ and combining them with the feature functions in \refeq{eq:discr} in order to obtain samples, say $m_b(\bs{s})$ of the sought material property field. Such samples can be generated as described in Equations (\ref{eq:ysamp}), (\ref{eq:xsamp}) earlier.
Estimates of the posterior mean $ \mathbb{E}[{m}(\ve s) | \hat{\bs{u}}, \hat{\bs{R}}]$ and variance $\mathrm{Var}[ m(\ve s) | \hat{\bs{u}}, \hat{\bs{R}}  ]$ can be obtained as follows:
\begin{align}
    \label{eqn:mean}
    \mathbb{E}[{m}(\ve s) | \hat{\bs{u}}, \hat{\bs{R}}] &= \frac{1}{B} \sum_{b=1}^B m_{b}(\ve s) \quad \mathrm{and} \\
    \mathrm{Var}[ m(\ve s) | \hat{\bs{u}}, \hat{\bs{R}}  ] &= \frac{1}{B } \sum_{b=1}^B \left( m_{b}(\ve s) - \hat{m}(\ve s) \right)^2. \label{eqn:variance}
\end{align}
Furthermore, we report in the subsequent numerical illustrations $95\%$ credible intervals which are obtained with the help of $2.5\%$ and $97.5\%$ posterior quantiles at each spatial location $\bs{s} \in \Omega$: 
\begin{align}
   Q_{0.025} (m(\ve s)) 
   &= \mathrm{quantile}(m_b(\ve s), 0.025) \quad \mathrm{and} \label{eqn:lower_bound}\\
    Q_{0.975} (m(\ve s)) 
    &= \mathrm{quantile}(m_b(\ve s), 0.975),  \label{eqn:upper_bound}
\end{align}
respectively. 
The illustrations contained in section \ref{sec:num_exp} were based on  $B= 1000$ posterior samples. An algorithmic summary of the steps described above can be found in Algorithm \ref{alg:mean_var}.

\begin{algorithm}[t]
\caption{Material field $m(\bs{s})$ posterior estimates}\label{alg:mean_var}
\begin{algorithmic}
\State Select $B$ \Comment{Number of samples}
\For{$b$ in $B$}
\State  $\ve y_b \sim q_{\bs{\xi}}(\ve y | ~ \ve \mu_{\ve y}, \ve S_{\ve y})$  \Comment{see \refeq{eq:ysamp}}
\State $\bs{\mu}_{b,x} \leftarrow NN(\ve y_b)$ 
\State $\ve x_{b} \sim q(\bs{\mu}_{b,x}, \ve S_{\ve x})$ \Comment{see \refeq{eq:xsamp}}
\State $m_{b}(\ve s) \leftarrow \sum \ve x_{b} \ve \eta^{\ve x}(\ve s)$ \Comment{see \refeq{eq:discr}}
\EndFor
\State Estimate $\mathbb{E}[{m}(\ve s) | \hat{\bs{u}}, \hat{\bs{R}}]$, $Var[{m}(\ve s) | \hat{\bs{u}}, \hat{\bs{R}}]$ 
\Comment{see \refeq{eqn:mean} / \refeq{eqn:variance}}
\State $Q_{0.025} (m(\ve s)) \leftarrow quantile(m_b(\ve s), 0.025)$ \Comment{see \refeq{eqn:lower_bound}}
\State $Q_{0.975} (m(\ve s)) \leftarrow quantile(m_b(\ve s), 0.975)$ \Comment{see \refeq{eqn:upper_bound}}
\end{algorithmic}
\end{algorithm}

%% file: Sections/4_NumericalExperiments.tex
In this section, we showcase the effectiveness of our approach on two-dimensional problems employing synthetic displacement data for high-dimensional linear and nonlinear PDEs. Our objectives are:
\bi
\item to compare in terms of accuracy and efficiency with  classical inference schemes such as Hamiltonian Monte Carlo (HMC, \cite{bishop2006pattern}) and Stochastic Variational Inference (SVI, \cite{hoffman2013stochastic}) that employ the forward solver as a black-box for the solution of the Bayesian inverse problem (section \ref{subsec:comparison}).
\item to demonstrate the proposed method's capability to produce accurate results in the presence of various noise levels (section \ref{subsec:noise}),
\item to highlight its ability to produce solutions to problems for which a well-posed, forward model is unavailable, as is the case when essential boundary conditions are not prescribed (section \ref{subsec:Dirichlet}),
\item to elucidate the proposed method's ability to handle seamlessly linear and nonlinear problems (section \ref{subsec:nonlinear}) with no alterations in the algorithmic framework and without any noticeable computational overhead.

\ei

\subsection{Problem setup and implementation details} \label{subsec:problem_setup}
\begin{figure}
    \centering
    \subfigure[]{\includegraphics{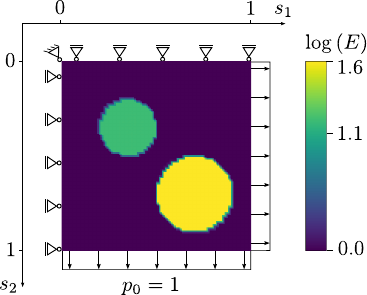}}
    \hspace{2cm}
    \subfigure[]{\includegraphics{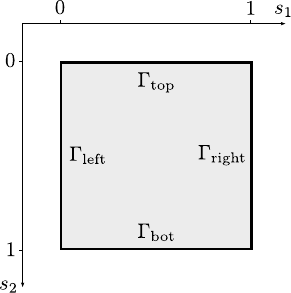}}
    \caption{(a): Problem configuration with \person{Dirichlet} BCs on boundaries $\Gamma_\mathrm{top}$ and $\Gamma_\mathrm{left}$ and \person{Neumann} BCs on boundaries $\Gamma_\mathrm{bot}$ and $\Gamma_\mathrm{right}$, respectively. The domain shows the ground truth material property field $m$ (e.g., the logarithm of \person{Young}s modulus) with two circular inclusions. (b): Boundary notation convention.}
    \label{fig:problem}
\end{figure}

The experimental setup (Figure \ref{fig:problem}) is motivated by  applications in elastography, where we want to identify inclusions (e.g., tumors) in healthy tissue.
The governing equations of small-strain,  solid mechanics described in Equations (\ref{eqn:PDE}), (\ref{eqn:const_law}) were employed for the square domain $\Omega=[0,1]^2$ and for the boundary conditions shown in  Figure \ref{fig:problem}.
The constitutive laws employed are described in the subsequent sections.
Reference solutions of the forward problem were obtained with FEniCS \cite{alnaes2015fenics, logg2012automated} using a triangular mesh with a total of  15,625 nodal points. 
The displacements on $\hat{\ve u}_{ref}=\{ u_{i,ref}\}_{i=1}^{N_{\hat{u}}}$ on a regular  $32 \times 32$ grid were subsequently contaminated with additive Gaussian noise with variance $\tau^{-1}$ in order to produce synthetic data $\hat{\bs{u}}$ used for the solution of the inverse problem. 
We report the noise variance  $\tau^{-1}$  in terms of the Signal-to-Noise (SNR) ratio defined as:
\begin{equation}
    \label{eqn:SNR}
    SNR_\mathrm{dB} = 10 \cdot \log_{10} \sqrt{\tau \frac{1}{N_{\hat{u}} } \sum_{i=0}^{N_{\hat{u}}}  u^2_{i,ref}  },
\end{equation}
which, unless otherwise mentioned, it is taken equal to $30 \mathrm{\ dB}$. We note that higher SNR means less noise and vice versa.

The black-box forward solver employed for the solution of the Bayesian inverse problem using classical inference techniques was based on  FEniCS \cite{alnaes2015fenics} with a regular mesh consisting of triangles as described in the sequel. The derivatives were obtained via PyAdjoint \cite{mitusch2019dolfin}.

For the solution of the inverse problem using the proposed method, we employed feature functions associated with the aforementioned uniform, triangular mesh over the problem domain. In particular, the $\eta^{\ve x}_i$ used in \refeq{eq:discr} to represent the  property field $m$ were piece-wise constant over each triangular element resulting in $dim(\bx)=d_x=1922$.
The feature functions $\bs{\eta}^{\ve y}_i$ used in \refeq{eq:discr} to represent the  displacement field $\bs{u}$ were taken to be the usual FE shape functions (i.e. piece-wise linear) associated with this triangular mesh resulting in $dim(\by)=d_y=2048$. 
While many other possibilities exist, as discussed earlier, the advantage of the representation adopted for $m$ is that the entries of $\bx$ are associated with each triangular element's center point. 
As a result, one can introduce structured, spatial priors on $\bx$  as explained in the sequel. 

\begin{figure}
    \centering
    \includegraphics{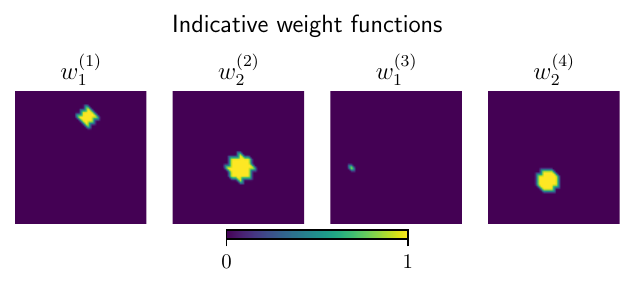}
    \caption{Illustration of 4 randomly selected weight functions $\ve w^{(j)}$ used in the algorithm. Note that for each weight function depicted are only the dimensions $w^{(j)}_1$ or $w^{(j)}_2$ which have non-zero values.}
    \label{fig:ws}
\end{figure}

The weight functions $\ve w(\ve s)$ employed in the weighted residuals, were expressed  with the same shape functions as $\ve u$ as follows:  
\begin{equation}
    \ve w(\ve z, \ve s) = \sum_{i=1}^{d_{\ve y}} z_i \bs{\eta}^{\ve y}_i(\ve s).
\end{equation}
Since each $\bs{\eta}^{\ve y}_i(\ve s)$, and therefore $z_i$, is associated with a nodal point of the aforementioned triangular mesh, we generated $N=24{,}576$ such weight functions $\bs{w}^{(j)}$ (from which $K$ were subsampled at each iteration of Algorithm \ref{alg:SVI}),  by drawing circles at uniformly sampled nodal locations and with random radii uniformly sampled in the interval $[ 0, 0.15 ]$. Subsequently, all the $z_i$ associated with nodal points falling within the circle were set equal to $1$, and the rest were equal to $0$. Figure \ref{fig:ws} depicts some indicative weight functions. 

An independent   prior $p(\bx,\by)=p(\by) p(\bx)$ was employed. For the representation of the displacement field i.e. $\by$ an uninformative, zero-mean,  Gaussian prior with isotropic variance $10^{16}$ was employed i.e. $p(\by)=\mathcal{N}(\by~ |~\bs{0}, 10^{16} \bs{I})$.
For the representation of the unknown material property field, i.e., $\bx$, a prior suitable for identifying inclusions was employed \cite{bardsley2013gaussian}. In particular, we denote with $\bs{J}_x=\bs{B~x}$ the vector of jumps between neighboring points of dimension $d_{jumps}$. We imposed a hierarchical prior on $\bs{J}_x$ that consists of:
\be
\label{eqn:jump_prior}
\begin{array}{c}
p(\bx| \bs{\theta})=\mathcal{N}( \bs{J}_x ~|\bs{0}, diag(\bs{\theta}^{-1})) \\
p(\bs{\theta}) =\prod_{j=1}^{d_{jumps}} Gamma(\theta_j | a_0,b_0)
\end{array}
\ee
The precision hyper-parameters $\bt$ promote sparsity in the number of jumps, i.e., piece-wise constant solutions \cite{bardsley2013gaussian}. Posed differently, amongst candidate solutions $\bx$, which fit the data equally well, the ones with the least total magnitude of the jumps between neighboring points are preferred. The values $a_0=b_0=10^{-8}$ were used in subsequent simulations \cite{bishop2013variational}. Inference of the associated hyper-parameters $\bt$ is discussed in Appendix \ref{app:jump_prior} and implies a negligible additional computational cost.

With regards to the approximate posterior $q_{\bs{\xi}}(\bx,\by)$ described in subsection \ref{subsec:approx_posterior} the following parameter values were employed $dim(\bs{\xi}_x)=15,946,000$ and $d_{\tilde{\ve x}}= d_{\tilde{\ve y}} =10$ which resulted in an overall dimension of the parameter vector $\bxi$ of $dim(\bs{\xi})=15,991,718$.  The parameters $\bs{\xi}_x$ correspond to a fully connected neural network with $3$ hidden layers and  $2000$ neurons  appearing  in \refeq{eqn:q_x}.

With regards to implementation details for the stochastic gradient ascent, we employed  ADAM \cite{kingma2014adam} with $\{\beta_1, \beta_2, lr \} = \{0.9, 0.99, 10^{-4}\}$. We employed $K=200$ (Algorithm \ref{alg:SVI}) weight functions per iteration as described in subsection \ref{subsec:ELBO} and $L = 10$ samples of  $(\ve x , \ve y)$ pairs from the approximate posterior for the associated Monte Carlo estimators (see \refeq{eqn:app_full_elbo}).  We found that increased efficiency can be achieved if one first updated  $q_{\bs{\xi}}(\ve y)$ separately on the observation data while not activating the virtual likelihood (which is equivalent to setting $\lambda = 0$ in \refeq{eqn:ELBO}).
We then train the joint approximate posterior until convergence 
with $\lambda = 10^7$. 
Our method was implemented in PyTorch \cite{NEURIPS2019_9015} and training was carried out on an Nvidia RTX 4090 GPU. The code can be found on the following repository upon publication \href{https://github.com/pkmtum/Weak-Neural-Variational-Inference}{https://github.com/pkmtum/Weak-Neural-Variational-Inference}.


\subsection{Accuracy and Efficiency comparison} \label{subsec:comparison}
In this section, we assess the proposed method against classical Bayesian inversion techniques that use the forward solver (and its adjoint for derivatives) as a black box.
The computational cost of our method is proportional to the number of weighted residual (and their gradient) evaluations. In order to make a fair comparison, we assume a first-order iterative solver is employed for the black-box version, which also relies on repeated evaluations of the weighted residuals and their derivatives. We note that each full update of such a solver requires computing all weighted residuals, whereas in our method $K$, such residuals are used for each SVI update.
Naturally, the total number of iterations in the former case would depend on the initial guess (which itself can be dependent on the inference scheme adopted) for the solution as well as the particulars of the update equations.
We discuss these aspects in detail in  Appendix \ref{app:comparison} where an equivalent cost in terms of weighted residual evaluations is established. We finally note that we did not use wall-clock time  as a cost metric as it depends on the software/implementation (vectorization, jit, parallelization) in relation to the available hardware (CPU / GPU).

We assume first  a linear, elastic constitutive law with Poisson's ratio $\nu=0.45$ and spatially variable Young's modulus $m$ given by (see Figure \ref{fig:problem}):
\be
    \label{eqn:youngsmod}
   \ln (m) = \begin{cases}
      1.6 & \textrm{if } (s_1-0.7)^2 + (s_2-0.7)^2 < 0.2^2 \\
      1.1 & \textrm{if } (s_1-0.35)^2 + (s_2-0.35)^2 < 0.15^2 \\
      0  & \textrm{else}.
    \end{cases}
\ee
The value selected for the first inclusion is based on  evidence  that  breast cancer tissue is approximately $5$ times stiffer than normal breast tissue \cite{cox2011remodeling, conklin2012stroma, wellman1999breast, mueller2010liver, masuzaki2007assessing}  whereas for the second inclusion on  the stiffness of a liver tumor being 3 times higher compared to healthy liver tissue \cite{mueller2010liver, masuzaki2007assessing}.

For the efficiency comparison only, we replace the jump prior in \refeq{eqn:jump_prior} with a simple \person{Gaussian} prior
\begin{equation}
    p(\ve x) = \mathcal{N}\left(\ve x | \mu = 0, \sigma= 2 \right),
\end{equation}
to ensure that we compare the (general) performance of the methods, not their ability to handle a specific prior.

\begin{figure}
    \centering
    \includegraphics{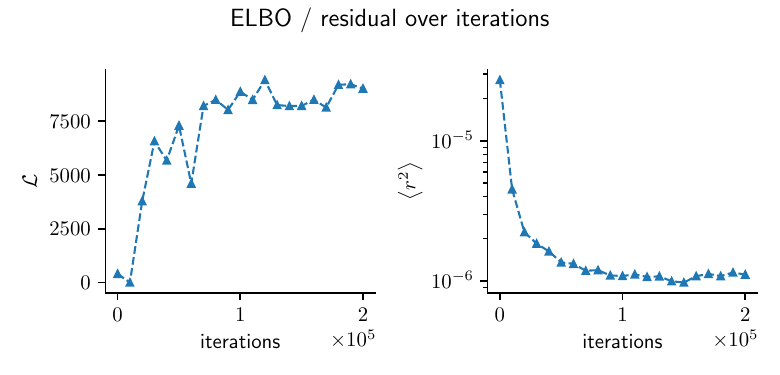}
    \caption{Convergence  of the proposed method. Depicted is the change of the ELBO $\mathcal{L}$ (left) and the expected squared residual $\left< r^2 \right>$ (right) over the iterations. The expected residuals are calculated using $50$ $\ve x$-$\ve y$ samples and all weight functions. 
    The fluctuations are due to the Monte Carlo estimates employed at each iteration.
    }
    \label{fig:elbo_res}
\end{figure}

\begin{figure}[!h]
    \centering
\includegraphics[width=.99\textwidth,height=0.72\textheight]{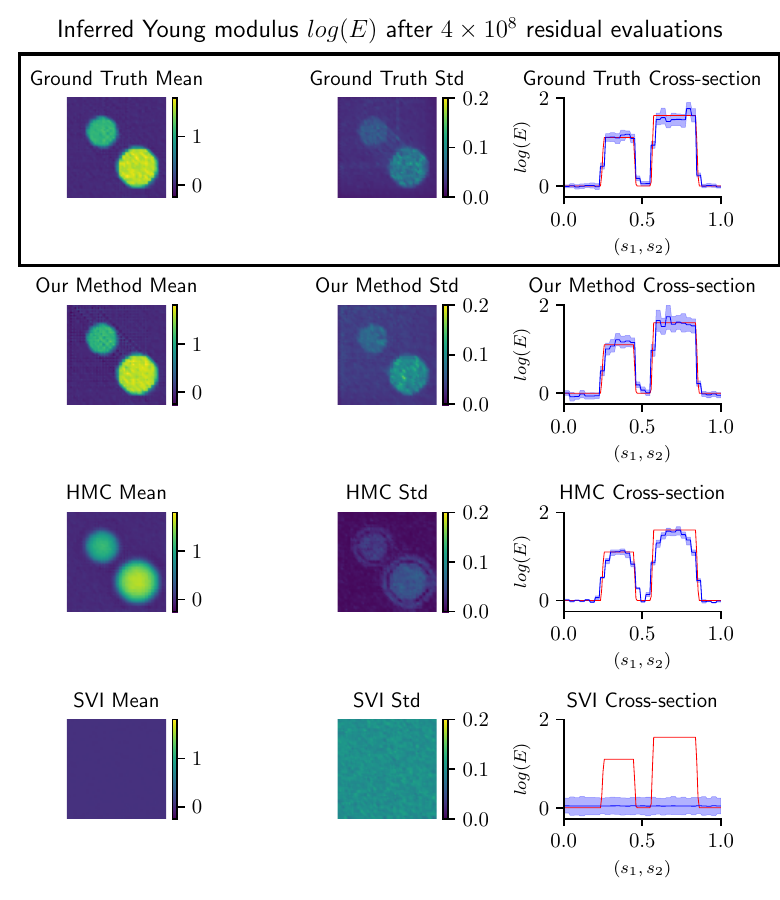}
    \caption{The three  columns contain posterior mean (first), posterior standard deviation (second) over the whole domain, and posterior estimates along the diagonal $s_1=s_2$ of the problem domain (i.e. posterior mean $\mu$ (blue line) and $95 \%$ credible intervals (blue shaded areas)). 
    The first row pertains to the  ground truth posterior obtained by black-box HMC run ($1.35 \times 10^{11}$ residual evaluations). The second row corresponds to  the posterior obtained by the proposed method  after $4 \times 10^8$ weighted residual evaluations. The third and fourth rows contain posterior estimates obtained with black-box HMC / SVI, respectively, and after the same number of residual evaluations as the proposed method.  The results were obtained with the same synthetic data contaminated by noise with SNR$=30$ dB. }
    \label{fig:comparison}
\end{figure}

The convergence of our method can be observed in Figure \ref{fig:elbo_res}, where the (rescaled) ELBO (left) and the expected square residual (right) over the iterations are depicted. In Figure \ref{fig:comparison}, we depict one-dimensional cross-sections along the diagonal $s_1 = s_2$ of the problem domain $\Omega$ of the inferred material field obtained using HMC, SVI, and the proposed method. The top row shows the ground truth posterior obtained by an HMC run with an equivalent cost of $1.35 \times 10^{11}$ residual evaluations ($300{,}000$ black-box forward solves). In contrast, our method (second row) converges to a posterior of similar quality after $4 \times 10^8$ weighted residual evaluations (i.e. approximately 3 orders of magnitude less). The slightly higher posterior uncertainty of our method could be attributed to the biased approximation of the posterior and the additional noise introduced by the virtual likelihood. For the same computational cost that it took for our method to converge, the black-box-solver-based HMC (third row) and SVI (fourth row) inference schemes produce estimates that deviate significantly from the ground truth (top row). 
These indicative plots showcase the superior efficiency of the proposed  method as compared to traditional, black-box-based  ones.

\subsection{Varying noise levels} \label{subsec:noise}
This subsection demonstrates the proposed method's performance under varying noise levels. 
In particular, we consider a linear elastic constitutive law with a spatially varying Young's modulus as in \refeq{eqn:youngsmod}. We consider three noise levels $SNR = \{25 \mathrm{ \ dB}, 30 \mathrm{ \ dB}, 35 \mathrm{ \ dB}\}$, defined as in \refeq{eqn:SNR}. Note that the  higher the SNR is, the  less noise is present in the data.

\begin{figure}
    \centering
    \includegraphics{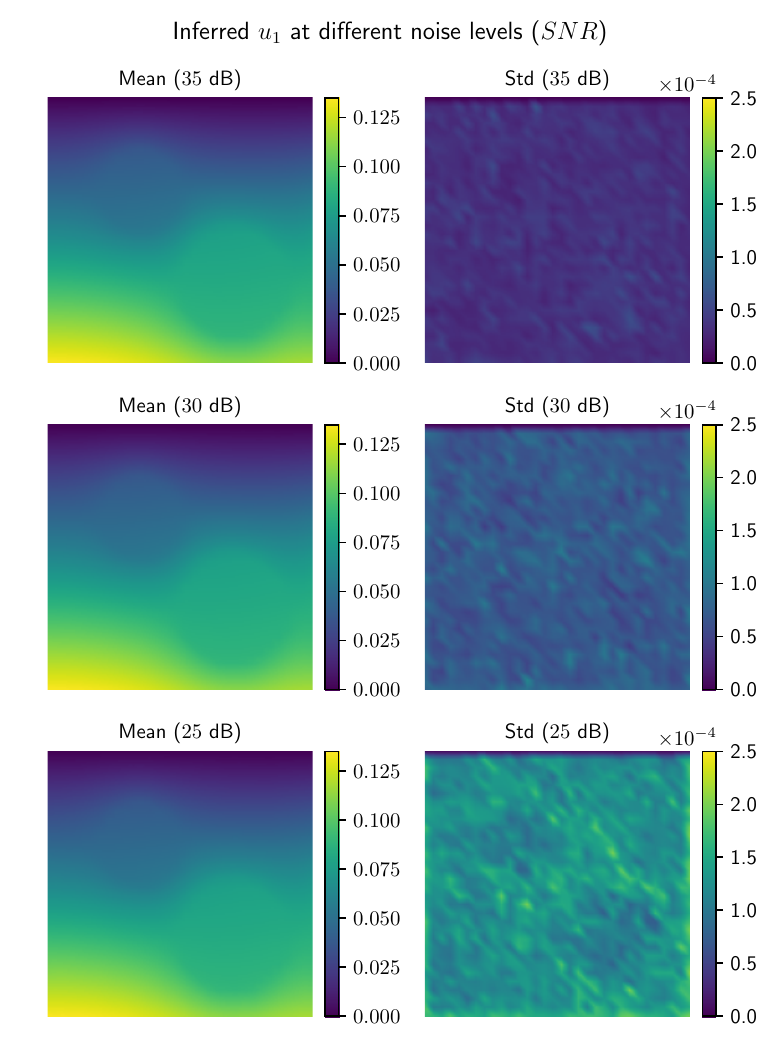}
    \caption{Posterior mean (left column) and standard deviation (right column) for the displacement field $u_1$ for SNR$= \{25 \mathrm{ \ dB}, 30 \mathrm{ \ dB}, 35 \mathrm{ \ dB}\}$. Note that the  higher the SNR is, the  less noise is present in the data.}
    \label{fig:u1_noise}
\end{figure}

\begin{figure}
    \centering
    \includegraphics{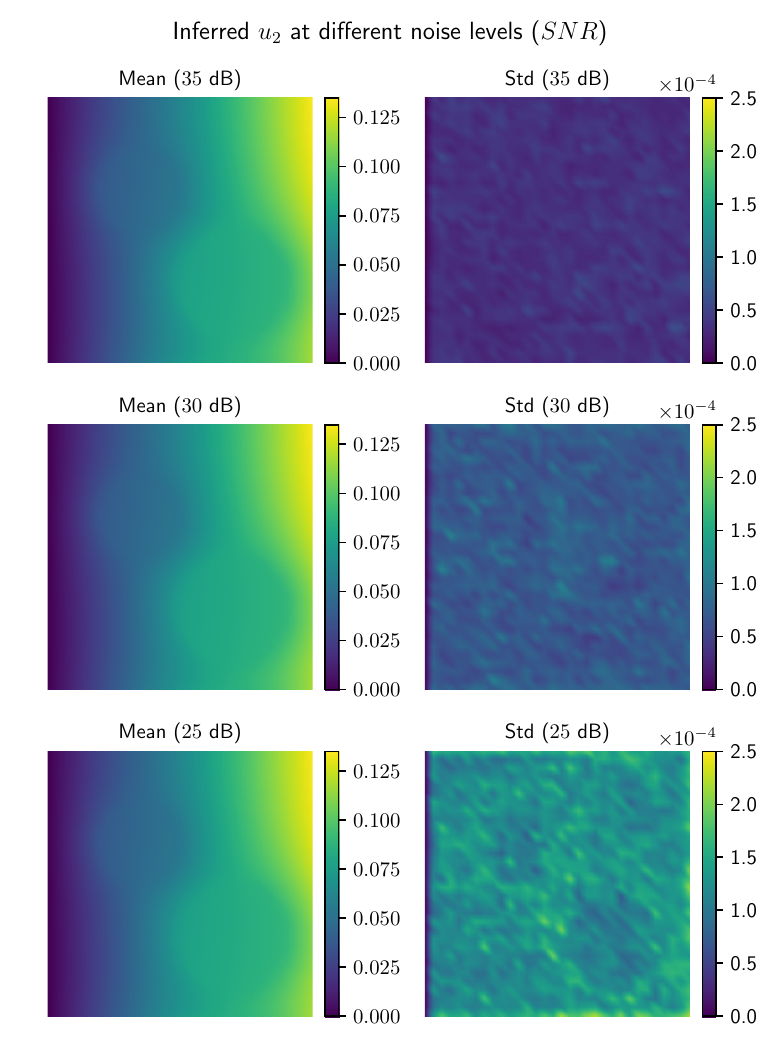}
    \caption{Posterior mean (left column) and standard deviation (right column) for the displacement field $u_2$ for SNR$= \{25 \mathrm{ \ dB}, 30 \mathrm{ \ dB}, 35 \mathrm{ \ dB}\}$. Note that the  higher the SNR is, the  less noise is present in the data.}
    \label{fig:u2_noise}
\end{figure}

\begin{figure}
    \centering
    \includegraphics{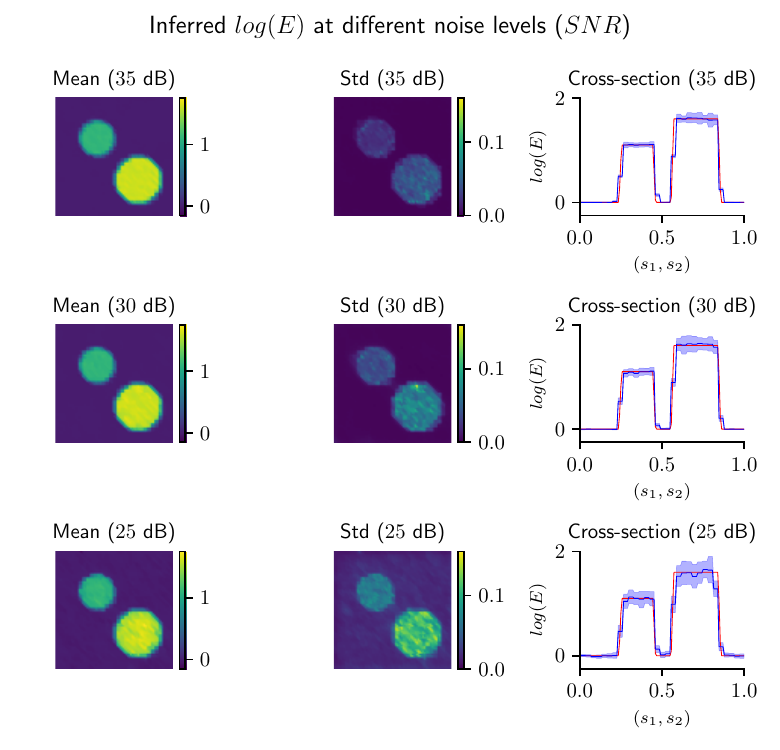}
    \caption{Posterior mean (left column), standard deviation (middle column) and posterior estimates along $s_1=s_2$ in the problem domain $\Omega$ (right column -  posterior mean $\mu$ (blue line)and  $95 \%$ credible intervals (blue shaded area)) of  the logarithm of  Young's modulus field for SNR$= \{25 \mathrm{ \ dB}, 30 \mathrm{ \ dB}, 35 \mathrm{ \ dB}\}$ (1. to 3. row), respectively. 
    Note that the  higher the SNR is, the  less noise is present in the data. 
    }
    \label{fig:E_noise}
\end{figure}

 Figures \ref{fig:u1_noise} and \ref{fig:u2_noise}  show the posterior mean and variance of the inferred displacement field (obtained through $\by$)  for $SNR=\{25 \mathrm{ \ dB}, 30 \mathrm{ \ dB}, 35 \mathrm{ \ dB}\}$, respectively. While the predicted means of the displacement fields $\ve u$ are nearly identical for all noise levels, their standard deviations increase with decreasing $SNR$ as one would expect. The inferred material field $m$ (obtained through $\bx$) and for the three SNRs is  shown in Figure \ref{fig:E_noise}. The respective posterior  means are almost identical and very close to the ground truth (Figure \ref{fig:problem}), 
 while as one would expect, the posterior variance increases when more noise is present in the data, i.e., for lower   $SNR$.
 Furthermore, we notice  that with increasing noise, the mean of the inclusions appears less homogeneous, and the edges are more diffuse. This is more clearly observed in the third column, which depicts posterior estimates along the diagonal $s_1 = s_2$ of the problem domain $\Omega$. 
 Nevertheless, we note that the credible intervals largely envelop the ground truth.

\subsection{Without Dirichlet boundary conditions} \label{subsec:Dirichlet}
As mentioned earlier, a necessary condition of traditional solution strategies for inverse problems is the availability of a well-posed forward problem and associated numerical solver. 
This has obfuscated the use of model-based elastography in clinical practice as the development of such solvers, apart from cost, requires particular knowledge and modeling skills. In this section, we aim to show that the proposed framework can produce accurate inverse problem solutions  without such a well-posed forward problem as would be the case when \person{Dirichlet} boundary conditions on $\Gamma_\mathrm{left}$ and $\Gamma_\mathrm{top}$ are unavailable (Figure \ref{fig:problem}). This missing information renders the forward problem ill-posed, constituting  classical schemes relying on \textit{black-box} solvers unusable. In contrast, our method treats the solution field of the PDE (i.e., the displacements in our case) as a random variable, irrespective of whether it pertains to the boundary or the interior of the problem domain, and infers it from the noisy observations and the weighted residuals.

\begin{figure}
    \centering
    \includegraphics{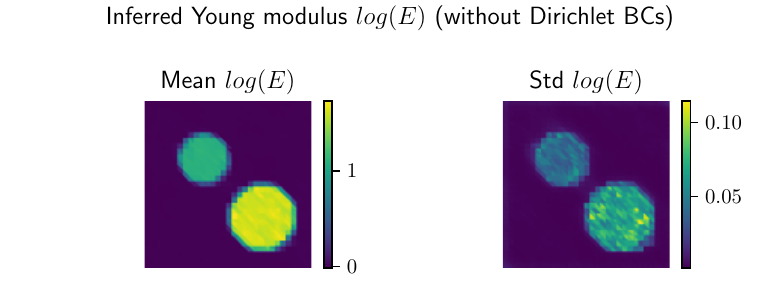}
    \caption{Inferred logarithmic \person{Young}s modulus mean and standard deviation fields when the \person{Dirichlet} boundary conditions are not given explicitly for SNR$=30$ dB.}
    \label{fig:noDirichlet}
\end{figure}

\begin{figure}
    \centering
    \includegraphics{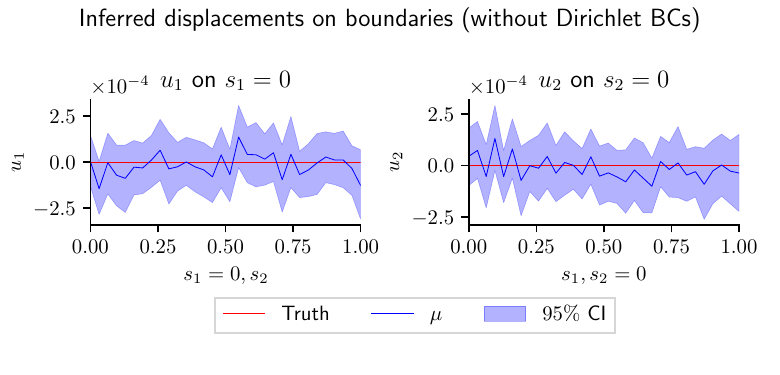}
    \caption{Inferred displacements on the \person{Dirichlet} boundaries when the \person{Dirichlet} boundary conditions are not given explicitly for SNR$=30$ dB. Depicted are the inferred mean $\mu$ (blue line) and $95 \%$ credible intervals (blue shaded areas) and the ground truth (red line).
    }
    \label{fig:boundary}
\end{figure}

The posterior mean and standard deviation of the unknown material field $m$ are depicted in Figure \ref{fig:noDirichlet} for $SNR=30dB$, which should be compared with the results in the second row of Figure \ref{fig:E_noise} that were obtained assuming given Dirichlet boundary conditions.   
The agreement is good in terms of the mean and the ground truth, but the posterior variance is at times larger due to the additional uncertainty.
In Figure \ref{fig:boundary}, we depict slices of the  inferred displacements along the aforementioned  boundaries $\Gamma_\mathrm{left}$ and $\Gamma_\mathrm{top}$.
One observes that these approximate the ground truth values which are always enveloped by  the posterior credible intervals. 
In addition, the computational effort associated with this case was identical to that for the case where the \person{Dirichlet} BCs were prescribed. 

\subsection{Comparison for non-linear material} \label{subsec:nonlinear}
Several studies have shown that human tissue exhibits non-linear material behavior \cite{oberai_linear_2009, o2009measurement}. To this end, we consider a nonlinear, Neo-Hookean constitutive law of the form \cite{ogden1997non}: 
\begin{equation}
    \sigma = 2 C J^{-1} (F F^T - I) + \left( 2 D \left( J - 1 \right) \right)  \otimes I,
\end{equation}
where $F = \left(I + \nabla u \right)$ denotes the deformation gradient, $J = \det\left( F \right)$ is its determinant, $I$ is the identity matrix, and $C, D$ are constitutive parameters. We assume that
\begin{equation}
    C(\ve s) = \frac{\tilde{E}}{4(1+\tilde{\nu})} \quad \mathrm{and \ } D(\ve s) = \frac{\tilde{E} \tilde{\nu}}{2 (1+\tilde{\nu}) (1-\tilde{\nu})},
\end{equation}
and for  $\tilde{\nu} = 0.45$, we  attempt to infer the spatially varying field $m(\ve s)=\log\left(\tilde{E}(\ve s)\right)$. The ground truth is assumed to be the field in  \refeq{eqn:youngsmod}. 

The presence of nonlinearity would necessitate the alteration of the forward solver in a black-box formulation, and each solution would generally imply an increased computational cost as compared to a linear model. 
In contrast, the only alteration in the proposed formulation pertains to incorporating the constitutive law in the weighted residual computations of \refeq{eqn:residual}. Its derivatives are automatically computed using automatic differentiation modules \cite{NEURIPS2019_9015}.

\begin{figure}
    \centering
    \includegraphics{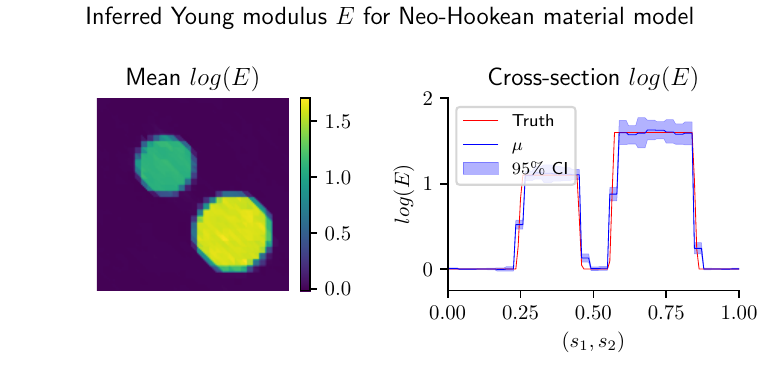}
    \caption{Inferred logarithmic material field $m$ cross-section (left) and mean field (right) for a non-linear Neo-Hook material law and SNR$= 30$ dB. In the cross-section depicted are the inferred mean $\mu$ (blue line) with $95 \%$ credibility intervals (blue regions) and the ground truth (red line).}
    \label{fig:NeoHook}
\end{figure}

The posterior mean of the inferred $m$  and the cross-sections obtained by the proposed method are shown in Figure \ref{fig:NeoHook}. The accuracy of the posterior estimates  is comparable to the previous cases where a linear model was employed. 
In all cases, the posterior mean is very close to the ground truth, which is largely enveloped by the credible intervals. 
More importantly, the computational cost to convergence was identical to that for a linear constitutive law. 
The fact that the inverse problem can be accommodated without any drastic changes in the solution procedure is a significant advantage to classical \textit{black-box} solver-based schemes, which also require  increased  computational efforts for such non-linear problems. 

%% file: Sections/5_Outlook.tex
We introduced a novel, data-driven method called Weak Neural Variational Inference (WNVI) for solving high-dimensional inverse problems based on PDEs as those arising in the context of model-based elastography.  This method solves the inverse problem within a Bayesian framework, i.e., it provides probabilistic estimates of the solution rather than point estimates, which is significant for applications requiring uncertainty quantification. 
The method employs a physics-aware probabilistic learning objective composed of real observations obtained from actual measurements and virtual observations derived from the weighted residuals of the governing PDE which  are used  as informational probes. As a result, information from the governing physical equations can be utilized {\em without } formulating or ever solving the forward problem. 
The formulation, apart from the usual unknowns, treats  the state variables of the physical model as latent variables that are inferred using Stochastic Variational Inference (SVI). The approximate posterior employed approximates the inverse map, from state variables to the unknowns,  with the help of neural networks. 
As demonstrated in the numerical experiments conducted, the proposed method exhibits several competitive advantages, which we summarize below:
\begin{enumerate}
    \item Each update in the SVI scheme requires evaluating a limited number of weighted residuals (i.e., integrals over the problem domain) and their derivatives, which can be readily obtained through automatic differentiation. Hence, the forward model {\em never} needs to be solved, leading to a highly efficient and flexible scheme. 
    \item Despite the reduced computational cost, the posterior estimates it produces exhibit similar accuracy with those obtained by an asymptotically exact Hamiltonian Monte Carlo (HMC) simulation, which employs the forward and adjoint solver as black-boxes. In practice, this simplifies the use of the method for medical practitioners, who generally do not have the expertise to set up a (nonlinear), discretized, forward model and its adjoint. This can significantly enhance the permeation of model-based in medical diagnostics and significantly simplify and expedite the computations required.
    \item It can successfully infer the unknowns not only for ill-posed inverse problems 
 but also for {\em ill-posed forward problems} as,  e.g., those arising when insufficient Dirichlet boundary conditions are prescribed. This addresses a common problem in elastography as boundary conditions are frequently ambiguous or completely unknown in \textit{in vivo} conditions.
    \item It  can handle nonlinear problems without methodological changes (apart from the integration module in calculating the weighted residuals) or incurring additional computational costs. In the case of elastography, nonlinear problems often arise from the nonlinear material behavior of human tissue.
\end{enumerate}
Moreover, the WNVI framework opens up numerous methodological and application possibilities, detailed below.
On the first front, the framework could benefit from a tempering scheme that progressively increases the precision $\lambda$ corresponding to the virtual observables or by employing alternative forms of virtual likelihoods to accelerate convergence. 
Secondly, given the role of the  weighting functions as information probes to the governing PDE, one can imagine that their informational content can vary significantly. This can be quantified in the probabilistic framework advocated  using the Evidence Lower Bound (ELBO). As a result, the method's efficiency could be greatly enhanced by adaptively adjusting the weighting functions/residuals to maximize informational gain. The potential speed-up by both previously proposed extensions could be vital in practice as it addresses computational
scalability issues and draw nearer to real-time, diagnostic capabilities even in cases where inexpensive, hand-held
ultrasound devices are used.

We note finally that the proposed method, as all model-based strategies,  assumes that the governing equations are  correct and reliable. In the context of the PDE-based models of continuum thermodynamics employed in this work, this is true for the conservation laws but not necessarily for  constitutive laws. While the method would always produce estimates of the unknown material parameters, those would be incorrect if the constitutive relation is invalid. In the context of model-based elastography this can lead to erroneous diagnostic decisions by the medical practitioners. 
It would be beneficial therefore to be able  to identify regions where the proposed constitutive equations are insufficient to model the material behavior. The framework advocated in this paper can be extended to incorporate different types of governing equations separately by employing different sets of virtual observables in order to achieve such a modification.

%% file: Sections/App_Integration.tex
\section{Numerical integration} \label{app:integration}
While general numerical integration techniques can be used for the integrals appearing in the weighted residuals in \refeq{eqn:residual}, we employed a method adapted to the particular feature functions used (see \refeq{eq:discr}) in the numerical illustrations.  In particular and given that these were associated with a uniform, triangle mesh, we used  closed form expressions over each triangle and summed over all of them in order to obtain the final result. 

%% file: Sections/App_ELBO.tex
\section{ELBO Terms} \label{app:ELBO}
We selected an uninformative Gaussian prior $p(\ve y) = \mathcal{N}\left(\ve y|~ \ve 0,~ \sigma^2 \ve I\right) $, where $\sigma^2 = 10^{16}$ (see section \ref{subsec:problem_setup}). We applied a prior on $\ve x$ that penalizes jumps between neighboring elements with penalty precision $\ve \theta$ as in \refeq{eqn:jump_prior}, which must also be inferred. 
 We adopt a mean-field approximation where the approximate posterior is of the form $q(\bx,\by,\bt)=q_{\bs{\xi}}(\bx,\by) q(\bt)$.
The approximate posterior $q_{\bs{\xi}}(\ve y, \ve x) = q(\ve x | \ve y) q(\ve y)$ consists of multivariate Gaussians with low-rank approximations of the covariance matrix and the mean of $q_{\bs{\xi}}(\ve x | \ve y)$ is represented  via a neural network (see subsection \ref{subsec:approx_posterior}). The approximate posterior $q(\ve \theta)$ is a Gamma distribution updates for which can be obtained in closed form.  Substituting the Monte Carlo approximation of the virtual likelihood term in \refeq{eqn:approx_virt_like} into ELBO formulation in \refeq{eqn:ELBO} yields in general
\begin{align}
     \mathcal{L}(\ve \xi) \approx& - \lambda \frac{N}{K} \sum_{k=1}^K \left< r^2_{\ve w^{j_k}} (\ve y, \ve x) \right>_{q_{\ve \xi}(\ve y, \ve x, \ve \theta)} \nonumber\\
     & -\frac{\tau}{2}  \sum_{i=1}^N \left< (\hat{\ve u}_i - \ve u_i(\by))^2 \right>_{q_{\ve \xi}(\ve y, \ve x, \ve \theta)} \nonumber\\
     &+ \left< \log p(\ve y , \ve x, \ve \theta) \right>_{q_{\ve \xi}(\ve y, \ve x, \ve \theta)} \nonumber \\
     & - \left< \log q_{\ve \xi}(\ve y, \ve x, \ve \theta) \right>_{q_{\ve \xi}(\ve y, \ve x, \ve \theta)}
     \mathrm{\ where \ } j_k \sim Cat\left(N, \frac{1}{N} \right). \label{eqn:app_elbo}
\end{align}
We note that the ELBO depends on the product $\lambda N$ rather than the individual terms which in the sequel \textbf{we represent  simply as $\lambda$}. Hence, in the sequel, the ELBO is approximated using the reparameterization trick \cite{kingma2013auto}. We can thus sample tuples $\{ \ve x_\ell, \ve y_\ell \}_{\ell=1}^L$ from the approximate posterior $q_{\bs{\xi}}(\ve x, \ve y)$ as described in subsection \ref{sec:elboalg} and use those calculate a Monte Carlo estimate of the expectations in \refeq{eqn:app_elbo}. Sampling from $q(\ve \theta)$ is not necessary, as we can update the posterior and calculate the needed expectations in closed form (see Appendix \ref{app:jump_prior}). Thus, \refeq{eqn:app_elbo} can be rewritten (up to an additive constant)
\begin{align}
     \mathcal{L}(\ve \xi) \approx& - \lambda \frac{1}{K L} \sum_{k=1}^K \sum_{\ell=1}^L r^2_{\ve w^{j_k}} (\ve y_\ell, \ve x_\ell) \nonumber\\
     & -\frac{\tau}{2 L}  \sum^{N\hat{u}}_{i=1} \sum_{\ell=1}^L  (\hat{\ve u}_i - \ve u_i(\by_\ell))^2 \nonumber\\
     & - \frac{1}{2 L} \sum_{\ell=1}^L \ve y_\ell^T  \left(\frac{1}{\sigma^2} \ve I \right) \ve y_\ell \nonumber \\
     & - \frac{1}{2 L} \sum_{\ell=1}^L \left( \ve L \ve x_\ell \right)^T diag(\left< \ve \theta \right>) \left( \ve L \ve x_\ell \right) \nonumber \\
     & + \log \left( \left< \ve \theta \right>^{a_0 - 1} \right) - b_0 \left< \ve \theta \right> \nonumber \\
     & - \frac{1}{2} \log \det \left( \ve S_y \right) \nonumber \\
     & - \frac{1}{2} \log \det \left( \ve S_x \right) \nonumber \\
     & + \sum_{i=1}^{I} \left(- a_{\theta_i} + \log b_{\theta_i} - \log \Gamma(a_{\theta_i}) - (1-a_{\theta_i}) \psi(a_{\theta_i}) \right), \label{eqn:app_full_elbo}
\end{align}
where we get $\left< \ve \theta \right > = \frac{a_{\theta_i}}{b_{\theta_i}}$ with \refeq{eqn:jump_update}. The third term can be ignored since we employ a large $\sigma$ in the uninformative prior of $\ve y$. As discussed in Appendix \ref{app:jump_prior}, we can iterate between first updating the approximate posterior $q(\ve \theta)$ in closed form and then updating the approximate posterior $q(\ve x, \ve y)$ using the derivatives of the ELBO in \refeq{eqn:app_full_elbo} obtained via automatic differentiation.

%% file: Sections/App_NN.tex
\section{Neural Network details} \label{app:NN}
The conditional mean $\bs{\mu}_{x, \bs{\xi}_x}$ of the approximate conditional posterior $q_{\bs{\xi}}(\ve x | \ve y)$ in \refeq{eqn:q_x} is parametrized by a neural network with tunable parameters $\bs{\xi}_x$. While virtually any neural network architecture is viable (e.g., CNN, deepONet, Deep Neural Network), we select a fully connected feed-forward neural network. For all our experiments we take the $d_{\ve y}$-dimensional input vector $\ve y$ and pass them through three sequential hidden layers with $2000$ neurons each and apply a \texttt{SiLU} (or \texttt{Swish} \cite{ramachandran2017searching}) activation function given by
\begin{equation}
    SiLU(x) = x \sigma(x),
\end{equation}
where $\sigma(x)$ denotes the sigmoid function. The last hidden layer is connected to the $d_{\ve x}$-dimensional output layer, which yields the conditional mean $\bs{\mu}_x$. We do not claim this design choice to be optimal in hyperparameter selection or architecture. Further studies can be conducted.

%% file: Sections/App_Jump_penalty.tex
\section{Jump penalty prior} \label{app:jump_prior}
To find a solution field with constant inclusions in a constant background, we employ a prior on $\ve x$ that penalizes jumps between neighboring elements as given in \eqref{eqn:jump_prior}. Here, we model the prior by a normal distribution where the mean is zero (i.e., no jumps), and the penalty for all jumps are given by a precision vector $\ve \theta$. The jump penalties $\ve \theta$ have a Gamma hyperprior, and we thus have to infer them in our approximate posterior. Note that the normal prior and the Gamma hyperprior are \textit{conjugate}, and we get a Gamma approximate posterior as a natural choice. We can, thus, update the approximate posterior in closed form as
\begin{equation}
    \label{eqn:jump_update}
    a_{\theta_i} = a_0 + \frac{1}{2}, \quad \mathrm{and} \quad b_{\theta_i} = b_0 + \frac{1}{2} J_i^2,
\end{equation}
where $J_i = (\ve L \ve x)_i$ is the respective value of the jump. Since the jump $J_i$ is dependent on $\ve x$, we suggest iterating between
\begin{enumerate}
    \item Updating the approximate posterior $q(\ve \theta)$ according to \eqref{eqn:jump_update} given samples of $\ve x$, and
    \item Updating the approximate posterior $q(\ve y, \ve x)$ according to the SVI scheme introduced in section \ref{sec:method}, where we can calculate $\left< \ve \theta \right> = \frac{a_{\theta_i}}{b_{\theta_i}}$ in closed form and the log probability of the prior and hyperprior becomes
    \begin{align}
        \log \, p(\ve J) &= \left< log \, p(\ve J | \ve \theta) \right> + \left< log \, \frac{p(\ve \theta)}{q(\ve \theta)} \right> \\
        &= - \frac{1}{2} \left< \ve J^T diag(\ve \theta) \ve J \right> \\
        &= - \frac{1}{2} \left< \ve J^T \right> diag \left( \frac{a_{\theta_i}}{b_{\theta_i}} \right) \left< \ve J \right>.
    \end{align}
\end{enumerate}

%% file: Sections/App_ComparisonDetails.tex
\section{Comparison details} \label{app:comparison}
The comparison in subsection \ref{subsec:comparison} uses the number of weighted residual evaluations  as a metric for computational cost. Our method evaluated $200$ weight functions for $10$ $\ve x - \ve y$-tuples, yielding $2000$ residual evaluations per iteration. The method took $200,000$ iterations to converge, which amounts to  $4 \times 10^8$ residual evaluations.

To compare black-box solver-based methods to ours, we considered an iterative solver. This requires the calculation of all residuals ($2048$) over each  update of $\ve y$ until the desired tolerance (absolute tolerance of $10^{-6}$) is reached. Starting from an initial vector $\ve y = \ve 0$, this takes on average $4400$ updates using the Generalized minimal residual method (GMRES). In HMC, however, subsequent steps yield similar results, and we can use the converged solution from the previous step as the initial guess for the next step. This reduces the number of required updates to $220$. This assumption cannot be made for SVI, as we draw independent samples from the posterior. We thus have $450,560$ and $9,011,200$ weighted residual evaluation equivalents per black-box forward solve for HMC and SVI, respectively. We, therefore, compare the posteriors after $880$ and $45$ black-box forward solves for HMC and SVI to our approximate posterior, respectively.